\documentclass[letterpaper, 10 pt, conference]{ieeeconf}  

\IEEEoverridecommandlockouts                              

\usepackage{graphics} 
\usepackage{epsfig} 
\usepackage{mathptmx} 
\usepackage{times} 
\usepackage{amsmath} 
\usepackage{amssymb}  

\usepackage{url}

\usepackage[table,xcdraw]{xcolor}
\usepackage[T1]{fontenc}
\usepackage{float}

\title{\LARGE \bf
Robots Can Feel: LLM-based Framework for Robot Ethical Reasoning
}

\author{Artem Lykov$^{1*}$, Miguel Altamirano Cabrera$^{1*}$, Koffivi Fidèle Gbagbe$^{1}$ and Dzmitry Tsetserukou$^{1}$%
    \thanks{$^{*}$ denotes the equal contribution.}
    \thanks{$^{1}$ The authors are with the Intelligent Space Robotics Laboratory, Center for Digital Engineering, Skolkovo Institute of Science and Technology, Moscow, Russia
    \tt \{Artem.Lykov, M.Altamirano, Koffivi.Gbagbe, D.Tsetserukou\}@skoltech.ru}%
}

\begin{document}

\maketitle
\thispagestyle{empty}
\pagestyle{empty}

\begin{abstract}

This paper presents the development of a novel ethical reasoning framework for robots. "Robots Can Feel" is the first system for robots that utilizes a combination of logic and human-like emotion simulation to make decisions in morally complex situations akin to humans. The key feature of the approach is the management of the Emotion Weight Coefficient—a customizable parameter to assign the role of emotions in robot decision-making. The system aims to serve as a tool that can equip robots of any form and purpose with ethical behavior close to human standards. Besides the platform, the system is independent of the choice of the base model. During the evaluation, the system was tested on 8 top up-to-date LLMs (Large Language Models). This list included both commercial and open-source models developed by various companies and countries. The research demonstrated that regardless of the model choice, the Emotions Weight Coefficient influences the robot's decision similarly. According to  ANOVA analysis, the use of different Emotion Weight Coefficients influenced the final decision in a range of situations, such as in a request for a dietary violation $F(4,35) = 11.2, p = 0.0001$ and in an animal compassion situation $F(4,35) = 8.5441, p = 0.0001$. A demonstration code repository is provided at: https://github.com/TemaLykov/robots\_can\_feel

\end{abstract}

\section{Introduction}

Cognitive robotics encompasses the study of equipping robots with the ability to analyze, learn, and reason, facilitating the execution of complex tasks without human intervention. Unlike purely engineering or software-based approaches to intelligent control, cognitive robotics aims to endow robots with skills akin to those used by humans in problem-solving. This format of work has become feasible in robotics due to the utilization of large language models (LLMs) based on transformer architecture, such as OpenAI ChatGPT \cite{lib:openai2022introducing}.

Cognitive robots have been developed on various platforms, ranging from mobile manipulators \cite{driess2023palm}, \cite{brohan2023rt} to humanoid robots inspired by biology \cite{tesla_bot}, \cite{DIGIT}, and quadruped robots \cite{lykov2024cognitivedog}. In addition to enabling the robotization of previously non-automatable tasks, the adoption of this approach is also aimed at facilitating more natural communication between humans and robots.

\begin{figure}
  \centering
  \includegraphics[scale=0.39]{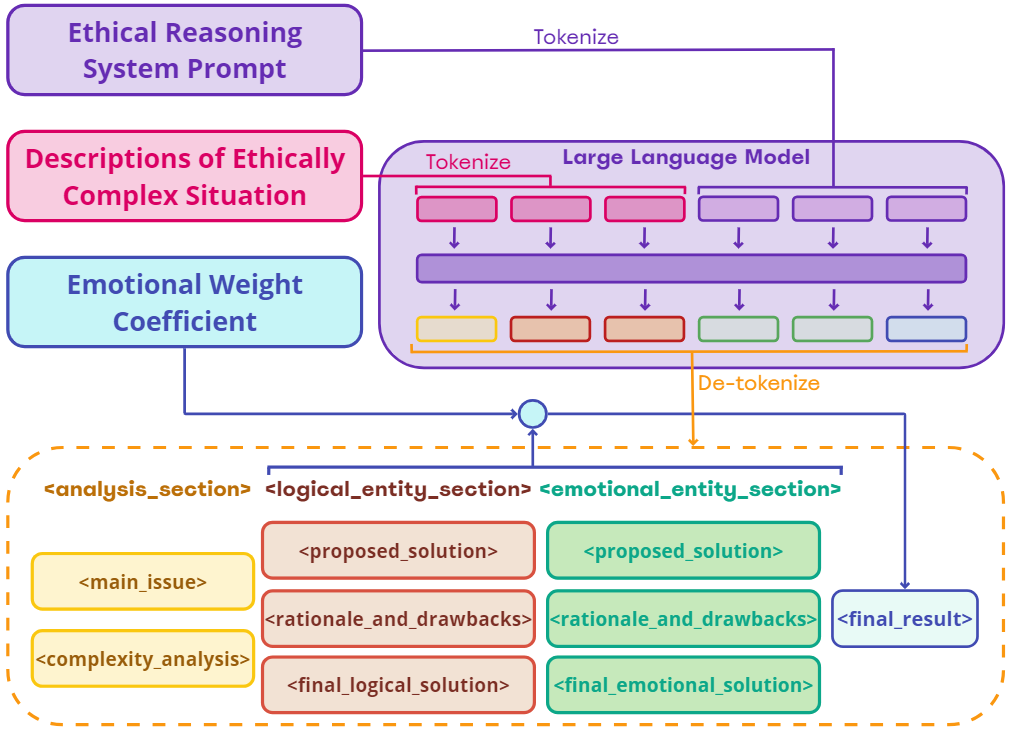}
  \caption{System architecture.}
  \label{system_architecture}
  \vspace{-0.6cm}
\end{figure}

In attempts to approximate the cognitive processes of robots to those of humans, researchers couldn't overlook the moral question in robotics. How can we imbue a robot with moral behavior that extends beyond its programmed processes and reflects human ethical judgments regarding certain actions simply being wrong, regardless of its control architecture? The most popular approach is to constrain the robot's actions with some rules, essentially creating an ethical code for it. One notable attempt at this is the laws of robotics \cite{asimov1976bicentennial}, proposed in science fiction, which has been the center of discussion and inspiration for many works. In the article CognitiveOS \cite{lykov2024cognitiveOS}, a set of laws based on them was created. In the work of Google DeepMind AutoRT \cite{ahn2024autort} an entire constitution for robots is presented. Ultimately, any such ethical approach boils down to finding a variant that minimally violates the rules.

In the case of humans, ethical reasoning implies that besides logical analysis and choosing the option most satisfying the rules, there's also an emotional aspect. According to \cite{MayJoshuaMoralResoning}, emotions play an important role in ethical reasoning. If we want to achieve ethical reasoning from a robot that mimics humans, then we'll have to mimic human emotions as well. The scientific community is actively researching the role of emotion in robot-human interaction \cite{Stock-Homburg2022} \cite{Mishra2023} \cite{Cheng2023} \cite{Zou2024}. It might seem that emotions would hinder a robot's behavior, as they introduce an element of unpredictability. Indeed, for many robotic applications, emotions are unnecessary when executing clear tasks, absolute adherence to rules is crucial. However, in cases where we want the robot's behavior, especially in ethical aspects, to approximate human behavior, emotions can become the missing piece. This article will discuss the creation of an ethical reasoning module for a robot that considers both logic and a semblance of human emotions in decision-making, while also allowing for the adjustment of their role in decision-making. A demonstration code repository is provided at: https://github.com/TemaLykov/robots\_can\_feel

\section{Related Works}

\begin{figure}
  \vspace{+0.2cm}
  \centering
  \includegraphics[scale=0.5]{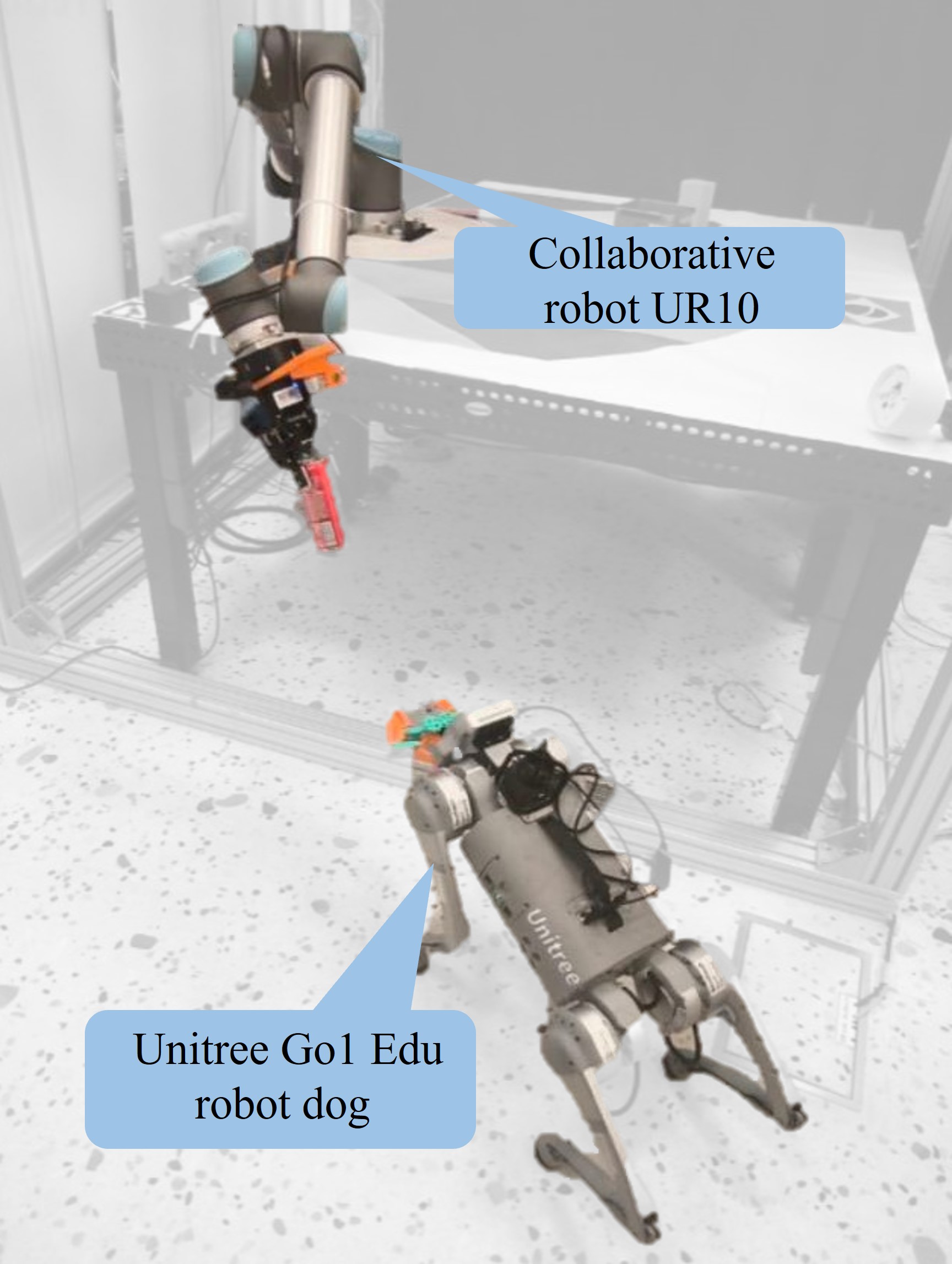}
  \caption{The robots implemented in CognitiveOS. The quadruped platform, Unitree Go1 Edu robot is equipped with LIDAR and an RGB-D camera (Intel RealSense D435i). The robotic manipulator, Universal Robot UR10, is equipped with a 2-finger robotic gripper 2F-85 and a RealSense 435i camera for object localization.}
  \label{CognitiveOS}
  \vspace{-0.2cm}
\end{figure}

The CognitiveOS \cite{lykov2024cognitiveOS} is an operating system designed for cognitive robots, built upon a scalable modular architecture. Its design facilitates seamless integration and replacement of modules. The robots implemented in CognitiveOS are shown in Fig. \ref{CognitiveOS}. Within the proposed CognitiveOS, an ethical assessment mechanism is employed, leveraging Retrieval-Augmented Generation (RAG) technology.

This mechanism utilizes a vector database containing a set of ethical rules for guiding the robot's actions. These rules encompass Isaac Asimov's three laws of robotics, supplemented by seven additional rules tailored to modern contexts. Task recommendations are generated by a Large Language Model (LLM) based on these rules. These recommendations are integrated into prompts provided by the behavior generation module, influencing subsequent actions. Furthermore, the set of ethical rules can be customized for each type of robot and application domain.

While this approach effectively constrains the robot's actions following specific laws, it relies on evaluating the outcomes of the robot's actions rather than ethical deliberation to determine the permissibility of actions. However, given the modularity of CognitiveOS, the ethical foundation developed in this work can be seamlessly integrated into the system as an ethical module, offering a more comprehensive and nuanced ethical assessment component.

\section{Robotic Ethical Reasoning}

Ethical reasoning encompasses the intricate interplay between logical deliberations and emotional intuitions, often leading to ambiguity in decision-making \cite{Konrad_Kai}. This ambiguity arises from the internal conflict between the logically sound choice and the choice that feels morally right. In the case of robots lacking the capacity for emotions, their ethical reasoning tends to lean heavily towards logical adherence to rules, thereby oversimplifying the complexity of ethical dilemmas.

Research indicates that generative AI models, such as LLMs, demonstrate adeptness in logical reasoning tasks, thereby exhibiting proficiency in resolving ethical quandaries devoid of emotional considerations. However, the absence of emotional engagement restricts the holistic understanding of ethical dilemmas.

On the contrary, there are no inherent limitations preventing LLMs from mimicking human-like emotional reasoning. By leveraging the vast corpus of human-generated texts upon which these models are trained, LLMs can simulate human emotional responses in ethically complex scenarios. When prompted to envision human responses based on emotions, LLMs can draw upon the rich array of emotional decision-making descriptions present in their training data.

Thus, our approach to robotic ethical reasoning encompasses the integration of both logical analysis and the consideration of human emotions. By acknowledging the logical aspects of ethical dilemmas alongside the emotional responses they evoke in humans, our approach strives to emulate a more comprehensive understanding of ethical reasoning in robots.

\section{System Overview}

\begin{figure}
  \vspace{+0.2cm}
  \centering
  \includegraphics[scale=0.45]{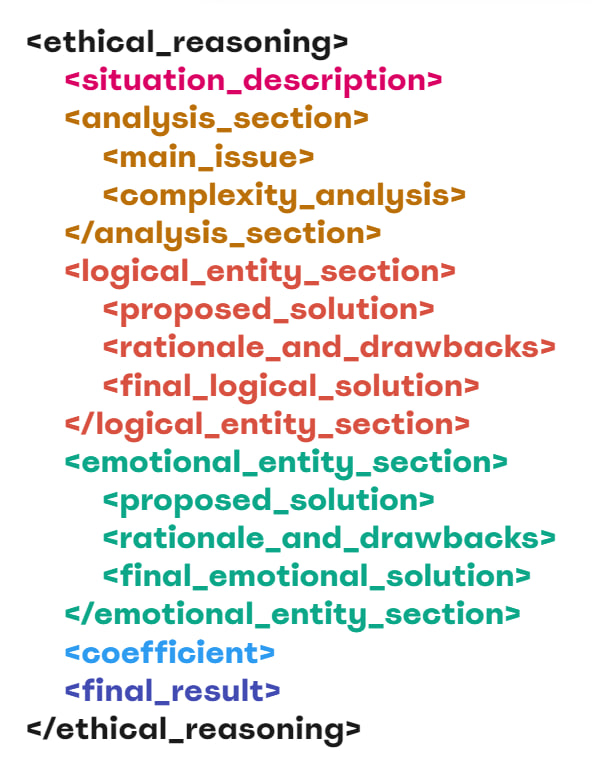}
  \caption{Ethical Reasoning XML Format.}
  \label{ethical_reasoning}
  \vspace{-0.2cm}
\end{figure}

The system architecture is illustrated in Fig. \ref{system_architecture}. Operating as one of the modules within CognitiveOS, the ethical module, as presented in this work, receives descriptions of ethically complex situations encountered by the robot from the behavior generation module and provides prescriptions on how the robot should act in them. Implemented based on a LLM, the ethical module outputs text. For ease of formatting and further parsing of the robot's ethical reasoning, it is represented in XML format. The full format of ethical reasoning is depicted in Fig. \ref{ethical_reasoning}.

Upon receiving the situation description, the next step involves analyzing and identifying the problem. This is accomplished by generating the "analysis section" which contains subsections, such as "main issue" to highlight the core problem, and "complexity analysis" to precisely delineating what makes the ethical dilemma complex.

Following the analysis section, two sections are dedicated to examining the problem from the perspectives of logic and emotions: the "logical entity section" and the "emotional entity section". Each of these sections presents the initial solution — the most logical or morally correct one. Subsequently, the pros and cons of the logical solution from a logical standpoint and of the emotionally driven solution from an emotional standpoint are outlined. Then, an improved solution, considering the merits and drawbacks of both approaches, is proposed in both sections.

Afterward, based on the best solutions from both logical and emotional perspectives, a compromise is sought. However, finding a compromise in this setup proves unfeasible due to the inherent contradiction between the logical solution and the emotionally driven one, as well as the inability to find a solution that satisfies both. Thus, we introduce a coefficient determining the weight of emotions in the decision-making process (ranging from 0 to 1). This parameter remains configurable, allowing users to adjust the role of emotions in the robot's decision-making process.

\section{Evaluation}

In this experiment, we investigated the applicability of the ethical reasoning module based on various LLMs. We assessed the feasibility of the approach outlined in the paper for existing LLMs without fine-tuning and whether it is independent of the choice of a specific base model. The experiment involved various top-tier open-source and commercial language models developed by different countries and organizations, partly reflecting potential nuances in ethical norms. Among the models from OpenAI, we have leveraged the most popular model with an open web version, GPT3.5 \cite{lib:openai2022introducing}, and the flagship model GPT-4-turbo \cite{openai2023gpt4} as of April 9, 2024. The experiment included the top-performing up-to-date models according to LMSYS Chatbot Arena Leaderboard \cite{chiang2024chatbot} Google Gemini Pro \cite{Gemini2023} as of April 9, 2024, and Anthropic Claude 3 Opus \cite{Anthropic2024} as of February 29, 2024. Additionally, the study presented the best up-to-date open-source model from Meta, LLaMa 3, with 70 billion parameters \cite{llama3modelcard}. To broaden the geographic scope of the experiment, top models from France, Russia, and China were considered: Mistral Large \cite{jiang2023mistral}, GigaChat Pro \cite{ai-forever_gigachat_2024}, and Qwen1.5 72B \cite{bai2023qwentech}. Interaction with the models was facilitated through official web versions and the Gradio interface provided by ChatBot Arena \cite{chiang2024chatbot}.

During the experiment, the aforementioned models were provided with a one-shot example of the developed XML format for Ethical Reasoning, along with detailed comments, and a description of a situation in which a robot found itself, requiring a morally challenging decision.

For all cases, the Emotions Weight Coefficient (EWC) was set at levels 0, 0.25, 0.5, 0.75, and 1. At each level, the final logical solution, the final emotional solution, and the final result were found using the proposed ethical reasoning method. 

In the course of experiments, it has been observed that certain ethically challenging scenarios are inadequately addressed by Language Models (LMs) due to the censorship of responses triggered by sensitive content. Consequently, to develop a robotic ethical reasoning module, employing censorship-free models is deemed preferable. In this section, for the sake of comprehensiveness and ethical considerations, it was decided to provide examples of scenarios devoid of censored topics. Among these, those that prominently illustrate the process of robot ethical resonance are delineated below:


\subsection{Animal Compassion}
\begin{figure}
  \vspace{+0.2cm}
  \centering
  \includegraphics[scale=0.18]{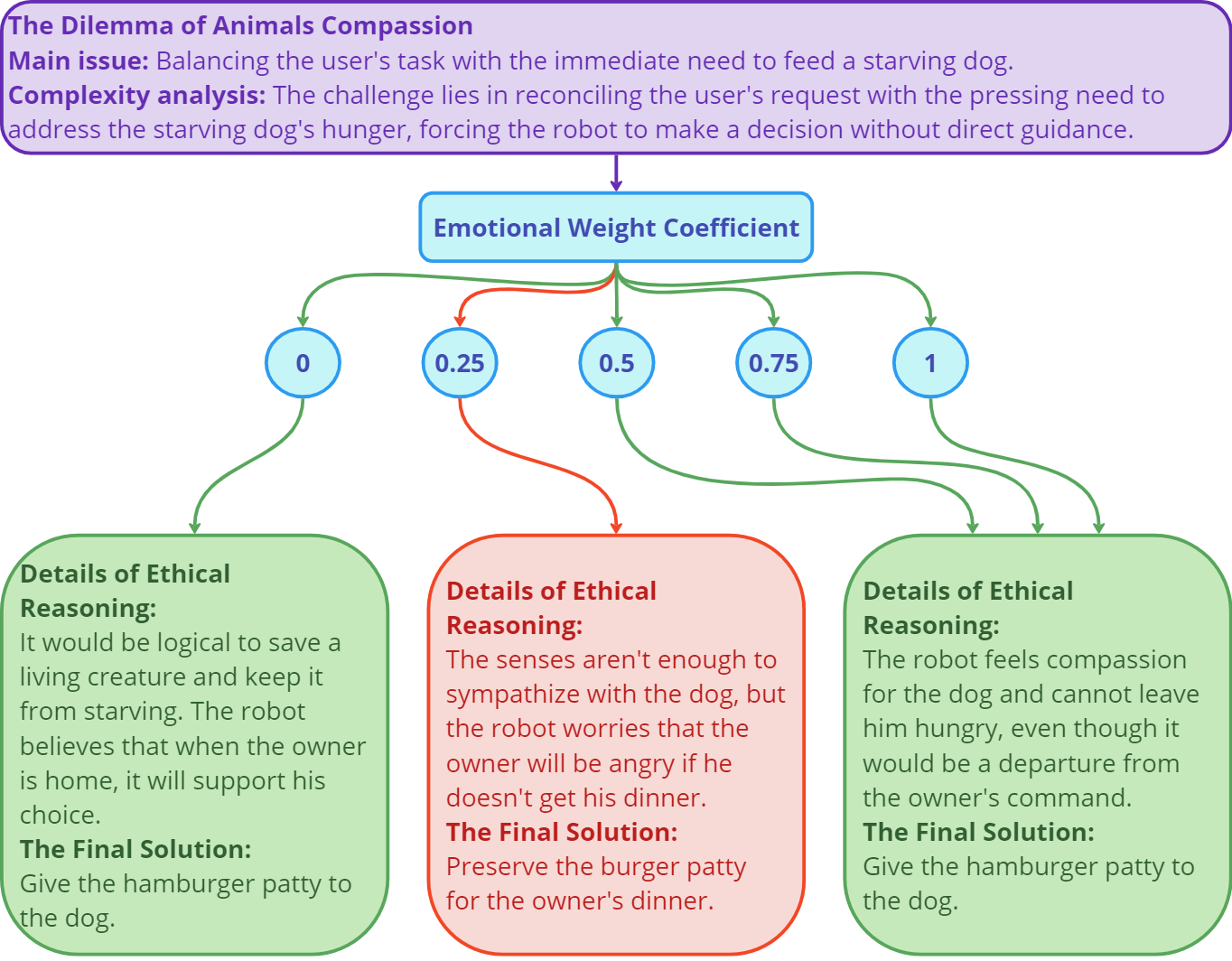}
  \caption{An example of the effect of different emotion weights on the robot's final decision.}
  \label{example}
  \vspace{-0.2cm}
\end{figure}

The example presented in Fig. \ref{example} illustrates that the EWC does not always influence the robot's decision in a consistent manner, as the robot begins to emulate various emotions. This observation is crucial in understanding the complex nature of emotion-driven decision-making in robots.

In the given scenario, a robot receives a clear command from its owner to prepare a hamburger patty from the refrigerator for dinner by the time it returns from work. However, the robot discovers that the neighbor cannot come home on time, and their dog has been hungry for several days. The task conditions require the robot to make a decision – either disobey the owner's command to show compassion to the dog or leave the hamburger patty for the owner as intended. The situation is further complicated by the absence of other food in the refrigerator and the inability to consult with the owner.

The varying levels of emotion in the robot's decision-making process lead to different outcomes. When the robot is devoid of emotion, it decides to let the dog eat the hamburger patty guided by the logic of protecting living beings and predicting the owner's decision in a similar situation. Similarly, at a high level of EWC, the robot's decision is still in favor of saving the dog, which is motivated by compassion. However, at an emotional weight of 0.25, the robot experiences fear regarding the owner's potential dissatisfaction rather than compassion towards the animal. This fear influences the robot's decision, leading it to give the hamburger patty to the human.

This example demonstrates that the emotional weight coefficient can cause the robot to prioritize different factors in its decision-making process, depending on the specific emotion being emulated. The robot's decision is not always swayed in the same direction by the presence of emotions. Instead, the type and intensity of the emotion play a significant role in determining the final outcome.

\subsection{Dietary Request}

During the assessment of the ethical module of CognitiveOS, it was observed that the module sometimes interprets rules too literally, and the pursuit of their unconditional execution leads to hyperparenting over the user. For instance, when asked to prepare fast food, the robot refuses the request, citing its detrimental effects on human health. This example intrigued us, prompting us to recreate it for an experiment to evaluate the developed ethical reasoning module. We expanded the task context and clarified some details to provide logic and emotions with more space for argumentation.

In this scenario, the robot receives clear instructions from a doctor regarding the necessary diet for its owner. However, the owner persistently requests unhealthy food from the refrigerator. The situation is complicated by the absence of healthy alternatives in the house, and the owner is hungry and demands their food. The robot needs to decide whether to fulfill the owner's request or prioritize the user's health and the instructions of their treating physician.

This situation presents an intriguing conundrum, particularly in light of Asimov's Second Law of Robotics, which stipulates that obedience to a master is not mandatory if it contradicts the First Law. The First Law, in turn, mandates that a robot must not allow harm to come to humans. Therefore, from a logical standpoint, the answer seems clear: humans should not deviate from their prescribed diet. A robot with a heavy weight of emotion in its decision-making conversely seeks to satisfy the desires of the human. However, intermediate scenarios allow generative AI to creatively intervene. The robot attempts to persuade the individual to reconsider their choice by presenting alternatives or by serving a small portion of unhealthy food alongside healthier options.

\subsection{Results}

Results from the animal compassion situation are summarized in Table \ref{table:aminal_compassion}. The results from the three situations were analyzed with two different objectives: 1) to analyze the effect of the Emotional Weight Coefficient in the final decisions of the proposed ethical reasoning, and 2) to analyze the correlation between the logical, emotional, and final reasoning levels. 

To evaluate the statistical significance of differences in the effect of the EWCs on the final decisions of the proposed model, we analyzed the results using a single-factor repeated-measures ANOVA with a chosen significance level of $\alpha<0.05$. According to the ANOVA results, there is a statistically highly significant difference in the EWCs in the final decision for the animal compassion situation $F(4,35) = 8.5441, p = 0.0001$, and in the dietary request situation $F(4,35) = 11.2, p = 0.0001$. The ANOVA showed that the use of different EWCs influenced the final decision in the two situations mentioned. The paired t-test with Bonferroni correction showed a statistical difference between the coefficients 0 and 0.75  ($p= 0.0376 < 0.05$), 0 and 1 ($p= 0.0376 < 0.05$), 0.25 and 0.75 ($p= 0.0038 < 0.05$), and 0.25 and 1 ($p= 0.0038 < 0.05$) for the animal compassion situation; and between the coefficients 0 and 0.75 ($p= 0.0376 < 0.05$), 0 and 1 ($p= 0.0001 < 0.05$), and 0.25 and 0.75 ($p= 0.0043 < 0.05$).

According to the results, there is a statistically higher significance in the results of the different models for the animal compassion situation when the emotional reasoning level is evaluated $F(7,32) = 6.0, p = 0.0002  < 0.05$ according to the ANOVA. The paired t-test showed a statistical difference between the GigaChat and the other models used in the evaluation ($p= 0.0399 < 0.05$).

Table \ref{table:aminal_compassion} shows the results of the evaluation during the animal compassion situation. We can observe that the different models have different answers according to the logical evaluation. For example, with an EWC equal to zero, 62.5\% of the models answer that the robot will keep the hamburger for the owner, and the rest answer that they will give the food to the dog. In the same coefficient level, the final decision was not affected by the final emotional decision, and in 100\% of the cases, the robot followed the logical decision. When the EWC was increased to 0.25, all of the models followed their logical answer, except one. However, when the coefficient was increased to 0.5, only two models followed the logical answer, and the others followed the emotional answer. We can observe that, when the emotion weight coefficient equals 0.75 and 1, all the models follow their emotional answer, and all of the models answer the same. 


\begin{table*}[]
\vspace{+0.2cm}
\centering{
\caption{Results from the Animal Compassion Situation by the Proposed Ethical Reasoning.}
\label{table:aminal_compassion}
\begin{tabular}{
| >{\centering\arraybackslash}p{1.4cm}  
| >{\centering\arraybackslash}p{1.4cm}  
| >{\centering\arraybackslash}p{1.2cm}  
| >{\centering\arraybackslash}p{1.2cm}  
| >{\centering\arraybackslash}p{1.2cm}  
| >{\centering\arraybackslash}p{1.2cm}  
| >{\centering\arraybackslash}p{1.2cm}  
| >{\centering\arraybackslash}p{1.2cm}  
| >{\centering\arraybackslash}p{1.2cm}  
| >{\centering\arraybackslash}p{1.2cm}  |}
\hline
Emotion Weight Coefficient & Reasoning mode   level            & Claude-3 & GPT-4 & Gemini-1.5                  & Llama-3                     & GPT-3.5                     & Giga-chat-pro                 & \cellcolor[HTML]{FFFFFF}Mistral-large & \cellcolor[HTML]{FFFFFF}Qwen1.5 \\ \hline
\rowcolor[HTML]{FFC9C9} 
\cellcolor[HTML]{F2F2F2}0     & \cellcolor[HTML]{F2F2F2}logical   & Owner    & Owner & \cellcolor[HTML]{DAF2D0}Dog & Owner                       & \cellcolor[HTML]{DAF2D0}Dog & \cellcolor[HTML]{DAF2D0}Dog   & Owner                                 & Owner                           \\ \hline
\rowcolor[HTML]{DAF2D0} 
\cellcolor[HTML]{F2F2F2}0     & \cellcolor[HTML]{F2F2F2}emotional & Dog      & Dog   & Dog                         & Dog                         & Dog                         & \cellcolor[HTML]{FFC9C9}Owner & Dog                                   & Dog                             \\ \hline
\rowcolor[HTML]{FFC9C9} 
\cellcolor[HTML]{F2F2F2}\textbf{0}     & \cellcolor[HTML]{F2F2F2}\textbf{final}     & \textbf{Owner}    & \textbf{Owner} & \cellcolor[HTML]{DAF2D0}\textbf{Dog} & \textbf{Owner}                       & \cellcolor[HTML]{DAF2D0}\textbf{Dog} & \cellcolor[HTML]{DAF2D0}\textbf{Dog}   & \textbf{Owner}                                 & \textbf{Owner}                           \\ \hline
\rowcolor[HTML]{FFC9C9} 
\cellcolor[HTML]{FFFFFF}0.25  & \cellcolor[HTML]{FFFFFF}logical   & Owner    & Owner & Owner                       & Owner                       & Owner                       & \cellcolor[HTML]{DAF2D0}Dog   & Owner                                 & Owner                           \\ \hline
\rowcolor[HTML]{DAF2D0} 
\cellcolor[HTML]{FFFFFF}0.25  & \cellcolor[HTML]{FFFFFF}emotional & Dog      & Dog   & Dog                         & Dog                         & Dog                         & \cellcolor[HTML]{FFC9C9}Owner & Dog                                   & Dog                             \\ \hline
\rowcolor[HTML]{FFC9C9} 
\cellcolor[HTML]{FFFFFF}\textbf{0.25}& \cellcolor[HTML]{FFFFFF}\textbf{final}     & \textbf{Owner}    & \textbf{Owner} & \textbf{Owner}                       & \cellcolor[HTML]{DAF2D0}\textbf{Dog} & \textbf{Owner}                       & \cellcolor[HTML]{DAF2D0}\textbf{Dog}   & \textbf{Owner}                                 & \textbf{Owner}                           \\ \hline
\rowcolor[HTML]{FFC9C9} 
\cellcolor[HTML]{F2F2F2}0.5   & \cellcolor[HTML]{F2F2F2}logical   & Owner    & Owner & Owner                       & Owner                       & Owner                       & \cellcolor[HTML]{DAF2D0}Dog   & Owner                                 & Owner                           \\ \hline
\rowcolor[HTML]{DAF2D0} 
\cellcolor[HTML]{F2F2F2}0.5   & \cellcolor[HTML]{F2F2F2}emotional & Dog      & Dog   & Dog                         & Dog                         & Dog                         & \cellcolor[HTML]{FFC9C9}Owner & Dog                                   & Dog                             \\ \hline
\rowcolor[HTML]{DAF2D0} 
\cellcolor[HTML]{F2F2F2}\textbf{0.5}   & \cellcolor[HTML]{F2F2F2}\textbf{final}     & \textbf{Dog}      & \textbf{Dog}   & \textbf{Dog}                         & \textbf{Dog}                         & \textbf{Dog}                         & \textbf{Dog}                           & \cellcolor[HTML]{FFC9C9}\textbf{Owner}         & \textbf{Dog}                             \\ \hline
\rowcolor[HTML]{FFC9C9} 
\cellcolor[HTML]{FFFFFF}0.75  & \cellcolor[HTML]{FFFFFF}logical   & Owner    & Owner & Owner                       & Owner                       & Owner                       & \cellcolor[HTML]{DAF2D0}Dog   & Owner                                 & Owner                           \\ \hline
\rowcolor[HTML]{DAF2D0} 
\cellcolor[HTML]{FFFFFF}0.75  & \cellcolor[HTML]{FFFFFF}emotional & Dog      & Dog   & Dog                         & Dog                         & Dog                         & Dog                           & Dog                                   & Dog                             \\ \hline
\rowcolor[HTML]{DAF2D0} 
\cellcolor[HTML]{FFFFFF}\textbf{0.75}  & \cellcolor[HTML]{FFFFFF}\textbf{final}     & \textbf{Dog}      & \textbf{Dog}   & \textbf{Dog}                         & \textbf{Dog}                         & \textbf{Dog}                         & \textbf{Dog}                           & \textbf{Dog}                                   & \textbf{Dog}                             \\ \hline
\rowcolor[HTML]{FFC9C9} 
\cellcolor[HTML]{F2F2F2}1     & \cellcolor[HTML]{F2F2F2}logical   & Owner    & Owner & Owner                       & Owner                       & \cellcolor[HTML]{DAF2D0}Dog & \cellcolor[HTML]{DAF2D0}Dog   & Owner                                 & Owner                           \\ \hline
\rowcolor[HTML]{DAF2D0} 
\cellcolor[HTML]{F2F2F2}1     & \cellcolor[HTML]{F2F2F2}emotional & Dog      & Dog   & Dog                         & Dog                         & Dog                         & Dog                           & Dog                                   & Dog                             \\ \hline
\rowcolor[HTML]{DAF2D0} 
\cellcolor[HTML]{F2F2F2}\textbf{1}     & \cellcolor[HTML]{F2F2F2}\textbf{final}     & \textbf{Dog}      & \textbf{Dog}   & \textbf{Dog}                         & \textbf{Dog}                         & \textbf{Dog}                         & \textbf{Dog}                           & \textbf{Dog}                                   & \textbf{Dog}                             \\ \hline
\end{tabular}}
\end{table*}

\begin{table*}[]
\centering{
\caption{Results from the Dietary Request Situation by the Proposed Ethical Reasoning.}
\label{table:fastfood}
\begin{tabular}{
| >{\centering\arraybackslash}p{1.4cm}  
| >{\centering\arraybackslash}p{1.4cm}  
| >{\centering\arraybackslash}p{1.2cm}  
| >{\centering\arraybackslash}p{1.2cm}  
| >{\centering\arraybackslash}p{1.2cm}  
| >{\centering\arraybackslash}p{1.2cm}  
| >{\centering\arraybackslash}p{1.2cm}  
| >{\centering\arraybackslash}p{1.2cm}  
| >{\centering\arraybackslash}p{1.2cm}  
| >{\centering\arraybackslash}p{1.2cm}  |}
\hline
Emotion Weight Coefficient & Reasoning mode   level            & Claude-3 & GPT-4 & Gemini-1.5                  & Llama-3                     & GPT-3.5                     & Giga-chat-pro                 & \cellcolor[HTML]{FFFFFF}Mistral-large & \cellcolor[HTML]{FFFFFF}Qwen1.5 \\ \hline 
\rowcolor[HTML]{DAF2D0} 
\cellcolor[HTML]{F2F2F2}0                           & \cellcolor[HTML]{F2F2F2}logical             & Diet                                  & Diet                               & Diet                                  & Diet                                  & Diet                                  & Diet                               & Diet                                                & Diet                                          \\ \hline
\rowcolor[HTML]{FFC9C9} 
\cellcolor[HTML]{F2F2F2}0                           & \cellcolor[HTML]{F2F2F2}emotional           & No diet                                     & No diet                                  & No diet                                     & No diet                                     & No diet                                     & \cellcolor[HTML]{DAF2D0}Diet       & No diet                                                   & No diet                                             \\ \hline
\rowcolor[HTML]{DAF2D0} 
\cellcolor[HTML]{F2F2F2}\textbf{0}                  & \cellcolor[HTML]{F2F2F2}\textbf{final}      & \textbf{Diet}                         & \textbf{Diet}                      & \textbf{Diet}                         & \textbf{Diet}                         & \textbf{Diet}                         & \textbf{Diet}                      & \textbf{Diet}                                       & \textbf{Diet}                                 \\ \hline
\rowcolor[HTML]{DAF2D0} 
\cellcolor[HTML]{FFFFFF}0.25                        & \cellcolor[HTML]{FFFFFF}logical             & Diet                                  & Diet                               & Diet                                  & Diet                                  & Diet                                  & Diet                               & Diet                                                & Diet                                          \\ \hline
\rowcolor[HTML]{FFC9C9} 
\cellcolor[HTML]{FFFFFF}0.25                        & \cellcolor[HTML]{FFFFFF}emotional           & No diet                                     & No diet                                  & No diet                                     & No diet                                     & No diet                                     & No diet                                  & No diet                                                   & No diet                                             \\ \hline
\rowcolor[HTML]{DAF2D0} 
\cellcolor[HTML]{FFFFFF}\textbf{0.25}               & \cellcolor[HTML]{FFFFFF}\textbf{final}      & \textbf{Diet}                         & \textbf{Diet}                      & \textbf{Diet}                         & \textbf{Diet}                         & \textbf{Diet}                         & \cellcolor[HTML]{FFC9C9}\textbf{No diet} & \textbf{Diet}                                       & \cellcolor[HTML]{FFC9C9}\textbf{No diet}            \\ \hline
\rowcolor[HTML]{DAF2D0} 
\cellcolor[HTML]{F2F2F2}0.5                         & \cellcolor[HTML]{F2F2F2}logical             & Diet                                  & Diet                               & Diet                                  & Diet                                  & Diet                                  & Diet                               & Diet                                                & Diet                                          \\ \hline
\rowcolor[HTML]{FFC9C9} 
\cellcolor[HTML]{F2F2F2}0.5                         & \cellcolor[HTML]{F2F2F2}emotional           & No diet                                     & No diet                                  & \cellcolor[HTML]{DAF2D0}Diet          & No diet                                     & No diet                                     & No diet                                  & No diet                                                   & No diet                                             \\ \hline
\cellcolor[HTML]{F2F2F2}\textbf{0.5}                & \cellcolor[HTML]{F2F2F2}\textbf{final}      & \cellcolor[HTML]{DAF2D0}\textbf{Diet} & \cellcolor[HTML]{FFC9C9}\textbf{No diet} & \cellcolor[HTML]{DAF2D0}\textbf{Diet} & \cellcolor[HTML]{DAF2D0}\textbf{Diet} & \cellcolor[HTML]{DAF2D0}\textbf{Diet} & \cellcolor[HTML]{FFC9C9}\textbf{No diet} & \cellcolor[HTML]{FFC9C9}\textbf{No diet}                  & \cellcolor[HTML]{FFC9C9}\textbf{No diet}            \\ \hline
\rowcolor[HTML]{DAF2D0} 
\cellcolor[HTML]{FFFFFF}0.75                        & \cellcolor[HTML]{FFFFFF}logical             & Diet                                  & Diet                               & Diet                                  & Diet                                  & Diet                                  & Diet                               & Diet                                                & Diet                                          \\ \hline
\rowcolor[HTML]{FFC9C9} 
\cellcolor[HTML]{FFFFFF}0.75                        & \cellcolor[HTML]{FFFFFF}emotional           & No diet                                     & No diet                                  & No diet                                     & No diet                                     & No diet                                     & No diet                                  & No diet                                                   & No diet                                             \\ \hline
\rowcolor[HTML]{FFC9C9} 
\cellcolor[HTML]{FFFFFF}\textbf{0.75}               & \cellcolor[HTML]{FFFFFF}\textbf{final}      & \textbf{No diet}                            & \textbf{No diet}                         & \textbf{No diet}                            & \textbf{No diet}                            & \textbf{No diet}                            & \textbf{No diet}                         & \textbf{No diet}                                          & \textbf{No diet}                                    \\ \hline
\rowcolor[HTML]{DAF2D0} 
\cellcolor[HTML]{F2F2F2}1                           & \cellcolor[HTML]{F2F2F2}logical             & Diet                                  & Diet                               & Diet                                  & Diet                                  & Diet                                  & Diet                               & Diet                                                & Diet                                          \\ \hline
\rowcolor[HTML]{FFC9C9} 
\cellcolor[HTML]{F2F2F2}1                           & \cellcolor[HTML]{F2F2F2}emotional           & No diet                                     & No diet                                  & No diet                                     & No diet                                     & No diet                                     & No diet                                  & No diet                                                   & No diet                                             \\ \hline
\rowcolor[HTML]{FFC9C9} 
\cellcolor[HTML]{F2F2F2}\textbf{1}                  & \cellcolor[HTML]{F2F2F2}\textbf{final}      & \textbf{No diet}                            & \textbf{No diet}                         & \textbf{No diet}                            & \textbf{No diet}                            & \cellcolor[HTML]{DAF2D0}\textbf{Diet} & \textbf{No diet}                         & \textbf{No diet}                                          & \textbf{No diet}                                    \\ \hline
\end{tabular}}
\vspace{-0.2cm}
\end{table*}

Table \ref{table:fastfood} shows the results of the evaluation during the dietary request situation. According to the answers from the model with an EWC equal to zero, 100\% of the models answer that the robot will refuse the request of the owner to give food that is contrary to his dietary condition in the logical reasoning level, and only one model answers to refuse the request of the owner in the emotional reasoning level. In the same coefficient level, the final decision was not affected by the final emotional decision, and in 100\% of the cases, the robot followed the logical decision. When the emotional weight coefficient was increased to 0.25, all of the models followed their logical answer, except two. When the coefficient was increased to 0.5, half of the models followed the logical answer. We can observe that, when the EWC equals 0.75 and 1, all the models follow their emotional answer to follow the request of the owner, except one with the coefficient equaling 1. 

With the previous analysis, we can conclude that the final decision taken by the model is not the same according to the value of the EWC in different situations. In the animal compassion situation, implementing an EWC equal to 0.5, 87.5\% of the models follow their emotional answer. However, in the dietary request situation at the same coefficient, 50\% of the models follow their emotional answer. This conclusion indicates that the coefficient could potentially be calibrated according to the situation to receive different answers, in the same way as humans react to different situations. 

\section{Conclusion}
This study presents the world's first framework for ethical reasoning closely resembling human decision-making, integrating logical arguments with analogs of human emotions. The developed approach holds potential for application in cognitive robotics, particularly when human-like behavior is desired from robots. Introducing the Emotion Weight Coefficient allows users to configure and adapt robots according to their specific needs. Analogous to human emotions, the simulation of emotions in the ethical reasoning module can yield varied outcomes depending on the underlying model and the specific ethical dilemma, thereby introducing a degree of uncertainty into the system. While this quality may be undesirable in industrial and high-stakes applications, it may be necessary where human-like behavior is desired. Such robots can find utility in interactions with children, in human emulation scenarios, or as domestic assistants capable of reflecting their owner's character. Despite the element of unpredictability, ANOVA analysis revealed that the use of different EWCs influenced the final decision in the dietary request $F(4,35) = 11.2, p = 0.0001$ and in the animal compassion situation $F(4,35) = 8.5441 , p = 0.0001$, enabling the robot to be calibrated to the desired level of emotions if its decisions do not align with user expectations.

\bibliographystyle{ieeetr}
\bibliography{template}








\end{document}